\documentclass{article}
\usepackage{spconf,amsmath,graphicx,threeparttable,booktabs,tabularx,subcaption,afterpage}
\usepackage{multirow}
\usepackage{color}

\newcommand{\eg}{e.\,g.,\ }
\newcommand{\ie}{i.\,e.,\ }

\title{Tailoring Generative Large Language Models for Specialised Emotion Recognition Tasks}
\title{Customising General Large Language Models for Specialised Emotion Recognition Tasks}

\name{Liyizhe~Peng$^1$, Zixing~Zhang$^1$*, Tao~Pang$^1$, Jing~Han$^2$*, Huan~Zhao$^1$*, Hao~Chen$^1$, Bj\"orn W.\ Schuller$^3$\thanks{$^*$ Corresponding authors. The work was funded by the National Science Foundation of China under Grant Number 62076092.}}
\address{$^1$ College of Computer Science and Electronic Engineering, Hunan University, China\\
$^2$ Department of Computer Science and Technology, University of Cambridge, UK\\ 
$^3$ GLAM, Department of Computing, Imperial College London, UK}

\begin{document}
\ninept
\maketitle
\begin{abstract}
The advent of large language models (LLMs) has gained tremendous attention over the past year. Previous studies have shown the astonishing performance of LLMs not only in other tasks but also in emotion recognition in terms of accuracy, universality, explanation, robustness, few/zero-shot learning, and others. Leveraging the capability of LLMs inevitably becomes an essential solution for emotion recognition. To this end, we further comprehensively investigate how LLMs perform in linguistic emotion recognition if we concentrate on this specific task. Specifically, we exemplify a publicly available and widely used LLM -- Chat General Language Model, and customise it for our target by using two different modal adaptation techniques, \ie deep prompt tuning and low-rank adaptation. 
The experimental results obtained on six widely used datasets present that the adapted LLM can easily outperform other state-of-the-art but specialised deep models. This indicates the strong transferability and feasibility of LLMs in the field of emotion recognition.

\end{abstract}
%


\section{Introduction}
Emotion recognition, a highly interdisciplinary research field spanning psychology, cognitive, and computer science, plays an increasingly important role in research related to human-computer interaction~\cite{DBLP:journals/cim/HanZS19}. Over the past decades, the domain of emotion recognition has undergone a profound transformation, thanks to the growing wealth of emotion datasets, enhanced computational capabilities, and continuous advancements in deep learning algorithms.

Recently, the emergence of large language models (LLMs), exemplified by ChatGPT and Claude, has ushered in a new era in the domain of emotion recognition. 
LLMs are typically pretrained on vast text corpora, showcasing their robust capabilities in various domains, including text generation and natural language understanding (NLU). Prior research has illuminated the remarkable capability of LLMs in the realm of emotion recognition, attaining commendable benchmarks in accuracy, universality, explanation, robustness, and few/zero-shot learning, among others~\cite{DBLP:journals/corr/abs-2308-11578}. However, it is imperative to address certain challenges. While few-shot learning can enhance model performance by providing limited demonstration examples within prompts, extending prompt length results in a quadratic escalation in inference computational costs. Furthermore, overly lengthy prompts risk truncation, as they may exceed maximum input limits, leading to diminished LLM output quality. As such, researchers and engineers are tasked with the endeavour to devise efficient methodologies for fine-tuning LLMs on domain-specific datasets.
The modal adaptation technique represents a training strategy meticulously crafted to further refine a pretrained model, with the principal objective of aligning the model's capabilities with specific tasks or domains. This methodology can facilitate the tailoring of pretrained LLMs to cater to particular downstream tasks, all while preserving their formidable language comprehension prowess.

To this end, we aim to shed some light on how LLMs perform if they are customised to the emotion recognition domain, and to find out whether they are competitive or better than a conventional deep model specifically designed for emotion recognition.  
For this purpose, we select a specific open-source LLM, \ie the Chat General Language Model, and employ two distinct model adaptation techniques: deep prompt tuning (P-Tuning v2) and low-rank adaptation (LoRA). Subsequently, we conduct a comprehensive comparative analysis, evaluating the performance of LLMs both pre- and post-adaptation on six emotional datasets. Furthermore, to provide a holistic perspective on the effectiveness and advancements achieved through these model adaptation approaches, we conducted comparative assessments with other state-of-the-art (SOTA) non-LLM-based studies. These comparative evaluations enable us to gauge the relative merits and contributions of different model adaptation strategies in the context of emotion recognition.
It is hoped that this work will bring more discussions in the field of emotion recognition, as the new era of general large models is coming.

\section{Related Work}
A substantial body of research has been dedicated to the domain of model adaptation for pre-trained LLMs. Within the spectrum of contemporary approaches, a widely-used and the most basic technique is known as full fine-tuning (FFT), necessitating the retraining of all model parameters. While FFT has proven effective in enhancing LLM performance, it demands substantial computational resources during training, incurring significant costs and rendering it increasingly impractical. In response to the need for reducing the computational burden, the research community has introduced numerous parameter-efficient fine-tuning (PEFT) methods~\cite{DBLP:journals/corr/abs-2303-15647}. These innovative methodologies entail the selective training of a limited subset of model parameters, either by modifying existing parameters or introducing novel ones into the model architecture.



The strategies for training partial model parameters involve adapting the characteristics of layer types or internal architecture within a network~\cite{DBLP:journals/corr/abs-2303-15647}. Methods involving the introduction of additional parameters can be broadly categorised into two groups: Adapter-like methods and Soft Prompts methods. Adapters introduce small fully-connected networks after Transformer sub-layers~\cite{DBLP:conf/emnlp/PfeifferRPKVRCG20}. 
Soft prompts can be trained for the input layer exclusively or for all layers within the model. Furthermore, reparametrisation-based PEFT methods leverage low-rank representations to minimise the number of trainable parameters.

In this paper, we choose two widely recognised model adaptation methods, namely P-Tuning v2 and LoRA. These two methods involve a limited number of trainable parameters, which ensures their computational resource demands remain within reasonable limits. Consequently, the adaptation training process can be feasibly conducted on consumer-grade graphics processing units (GPUs), rendering both P-Tuning v2 and LoRA highly practical choices.

\begin{table*}[t!]
\caption{Detailed information of the selected six datasets. \#sp., \#dia., \#total (test) utt., \#words/utt., \#classes denotes the number of distinct speakers, dialogues, utterances of the whole dataset and its test subset, words per utterance, and emotional classes.}
\vspace{-5pt}
\label{table:datasets}
\begin{threeparttable}
\centering
\resizebox{\linewidth}{!}{
\begin{tabular}
{p{1.5cm}<{\centering}p{1.2cm}<{\centering}p{1.2cm}<{\centering}p{1cm}<{\centering}p{1.5cm}<{\centering}p{0.6cm}<{\raggedleft}p{0.7cm}<{\raggedleft}p{2.0cm}<{\raggedleft}p{1.0cm}<{\raggedleft}p{3.5cm}<{\centering}}
\toprule
\textbf{dataset} & \textbf{language} & \textbf{modality} & \textbf{dialogue} & \textbf{data source} & \textbf{\#sp.} & \textbf{\#dia.} & \textbf{\#utt. total (test) } & \textbf{\#words/utt.} & \textbf{\#classes} \\
\midrule

SST & English & t & no & movie review & - & - & 11\,855 (2\,210) & - & 5 (negative, somewhat negative, neutral, positive, somewhat positive) \\

Friends & English & t & yes & Friends TV shows & - & 1\,000 & 14\,503 
 (2\,764) & 10.7 & 7 (neutral, joy, sadness, fear, anger, surprise, disgust)\\

Mastodon & English & t & yes & Mastodon & - & 505 & 2\,217 (1\,142) & - & 3 (positive, neutral, negative)\\

MOSI & English  & a, v, t & no & YouTube & 89 & - & 2\,199 (686) & 12.0 & 7 \{-3, -2, -1, 0, 1, 2, 3\}\\

CH-SIMS & Mandarin & a, v, t & no & movies, TVs, \& shows & 474 & - & 2\,281 (457) & 15.0 & 5 \{-1.0, -0.8\}\{ -0.6, -0.4, -0.2\} \{0.0\} \{0.2, 0.4, 0.6\}\{0.8, 1.0\}\\

M$^3$ED & Mandarin & a, v, t & yes & TV series & 626 & 990  & 24\,449 (4\,201) & 7.4 & 7 (happy, surp., sad, disgust, anger, fear, neut.)\\

\bottomrule
\end{tabular}}
\end{threeparttable}
\vspace{-10pt}
\end{table*}

\begin{figure}[t!] 
  \centering
  \begin{subfigure}{.6\linewidth}  
    \centering
    \includegraphics[width=\linewidth]{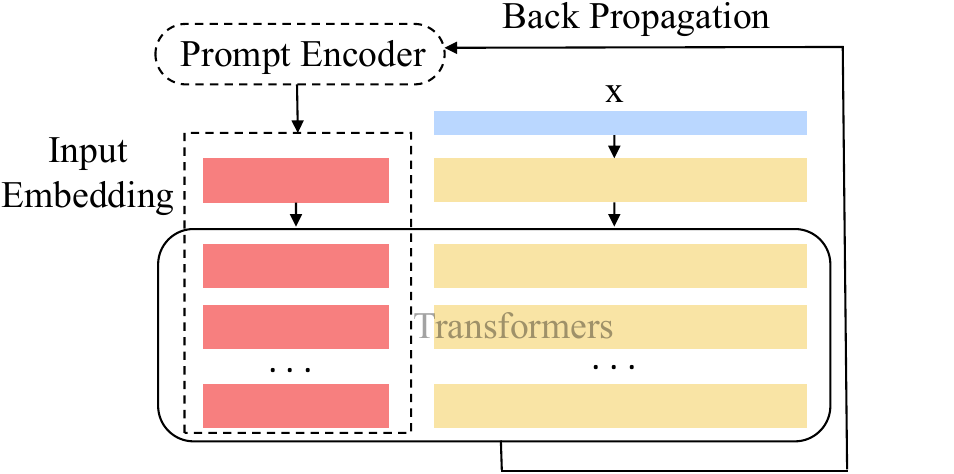}
    \caption{P-Tuning v2}
    \label{fig:P-Tuning}
  \end{subfigure}%
  \begin{subfigure}{.4\linewidth}  
    \centering
    \includegraphics[width=\linewidth]{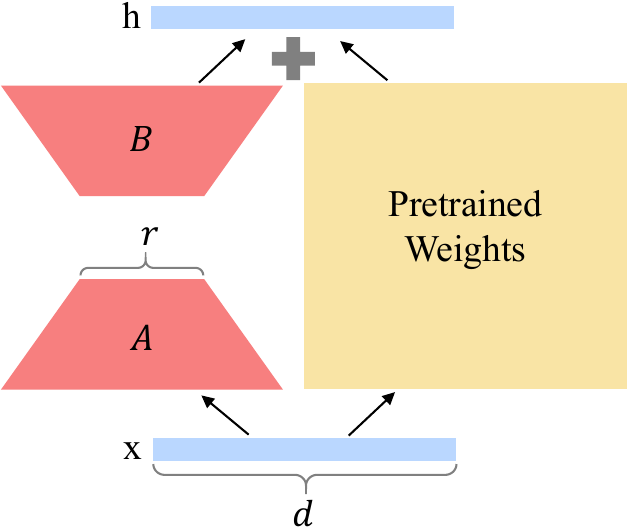}
    \caption{Low-Rank Adaptation}
    \label{fig:LoRA}
  \end{subfigure}
  
  \caption{Schematic representation of two model adaptation methods: P-Tuning v1 (a) and Low-Rank Adaption (b). Red blocks refer to the additional trainable parameters, while the yellow blocks represent the frozen parameters of the pre-trained model.}
  \label{fig:main}
  \vspace{-10pt}
\end{figure}

\section{Adaptation of Large Language Models}
In this section, we present an introduction to the selected LLM -- Chat General Language Model, alongside an explanation of the core principles underlying the two model adaptation techniques, namely P-Tuning and LoRA.
\vspace{-8pt}
\subsection{General Language Model}
General Language Model (GLM) is a general pre-training framework based on a novel autoregressive blank infilling objective and can be adapted to various NLU and natural language generation tasks~\cite{DBLP:conf/acl/DuQLDQY022}. GLM formulates NLU tasks as `cloze' questions that contain task descriptions, which can be answered by autoregressive generation. Remarkably, when operating with equivalent parameter counts and computational resources, GLM consistently outperforms BERT on the SuperGLUE benchmark, and excels beyond RoBERTa and BART when pretrained on corpora of comparable sizes~\cite{DBLP:conf/acl/DuQLDQY022}. Furthermore, GLM demonstrates superior performance to T5 in 
NLU and generation tasks, while utilising fewer parameters and data~\cite{DBLP:conf/acl/DuQLDQY022}. 

Chat General Language Model (ChatGLM), launched in March 2023 by the Tsinghua University KEG Laboratory and the Zhipu AI Company, is a GLM-based AI Chatbot. ChatGLM draws inspiration from ChatGPT to integrate code pre-training into the trillion-parameter base model GLM-130B~\cite{Zeng23-GLM}, 
achieving human intent alignment through techniques like supervised fine-tuning. It is worth mentioning that ChatGLM was pre-trained on both Chinese and English corpora, thus, it possesses bilingual capabilities. ChatGLM2, a second-generation model, is open-sourced in June 2023. It boasts enhanced performance, extended context capabilities, improved inference efficiency, and a more permissive open-source license. In our study, we opted to utilise the more lightweight model ChatGLM2-6B, which employs the same technology as ChatGLM2 but with a reduced parameter count of 6.2 billion. It requires a minimum of 13GB of GPU memory for inference when using FP16 precision, enabling the deployment of ChatGLM-6B on consumer-grade graphics cards.

\vspace{-10pt}
\subsection{P-Tuning}
P-Tuning (\textit{aka} Prompt Tuning) is a cost-effective model adaptation methodology, which freezes all model parameters and introduces additional parameters~\cite{DBLP:journals/corr/abs-2110-07602}. It can be viewed as an optimised and tailored implementation of deep prompt tuning, specifically designed for generation and knowledge probing tasks. Deep prompt tuning expands the capacity of continuous prompts, bridging the adaptation gap across various settings, with particular efficacy for smaller models and challenging tasks. Nevertheless, the P-Tuning method is subject to certain constraints due to its exclusive utilisation of continuous prompts within the input embedding sequence. This results in a limited number of trainable parameters, and the input embeddings have a relatively indirect impact on model predictions.

To overcome these challenges, the P-Tuning v2 technique incorporates continuous prompts into every layer of the model (shown in Fig.~\ref{fig:main} (a)), rather than solely in the input embedding sequence~\cite{DBLP:journals/corr/abs-2110-07602}. Such adjustment in P-tuning v2 introduces a greater number of tunable task-specific parameters (from 0.01\% to 0.1\%-3\%) to enhance task-specific capacity while maintaining parameter efficiency. Additionally, prompts added to deeper layers have a more direct impact on model predictions. Currently, P-Tuning v2 consistently achieves comparable performance to fine-tuning across a broad spectrum of model scales, ranging from 300M to 10B parameters~\cite{DBLP:journals/corr/abs-2110-07602}. It was frequently demonstrated to perform particularly well on challenging sequence tagging tasks, including extractive question answering and named entity recognition.

\vspace{-8pt}
\subsection{Low-Rank Adaptation}
Low-Rank Adaptation (LoRA) is one of the reparametrisation-based model adaptation methods~\cite{DBLP:conf/iclr/HuSWALWWC22} and is illustrated in Fig.~\ref{fig:main} (b). It freezes the pre-trained model weights and injects trainable low-rank decomposition matrices into each layer of the Transformer architecture,  
considerably reducing the number of trainable parameters for downstream tasks. Its inspiration comes from a statement 
that pre-trained language models have a lower ``intrinsic dimension'' and can still effectively learn despite being randomly projected into smaller subspaces~\cite{DBLP:conf/iclr/HuSWALWWC22}. Therefore, researchers hypothesised the updates to the weights also have a low ``intrinsic rank'' during adaptation and proposed LoRA methods. Compared to GPT-3's 175B parameters fine-tuned with Adam, LoRA can achieve a 10,000-fold reduction in the number of trainable parameters and a 3-fold decrease in GPU memory requirements~\cite{DBLP:conf/iclr/HuSWALWWC22}. In terms of effectiveness, LoRA matches or surpasses full fine-tuning for RoBERTa, DeBERTa, GPT-2, and GPT-3, even though it has fewer trainable parameters and a higher training throughput~\cite{DBLP:conf/iclr/HuSWALWWC22}. More importantly, it does not introduce any additional inference latency. 

\section{Experiments and Results}
\vspace{-5pt}
\subsection{Selected Datasets}\label{sec:datasets}

In this part, we present the six datasets utilised in our research. More detailed information is placed in Table 1, ranging from English to Chinese languages, from binary/ternary sentiment anlaysis to multi-class emotion classification. Since these datasets are publicly accessible, we can use them to verify the effectiveness of different model adaptation methods.

\textbf {SST}: 
The Stanford Sentiment Treebank is an English corpus containing fine-grained sentiment annotations for 11,855 individual sentences sourced from movie review data~\cite{socher2013recursive}. For the fine-grained task, each sentence is categorised into one of five sentiment classes. 
For the binary task, each sentence is simply classified as either positive or negative, with the neutral category excluded. 

\textbf {Friends}:
Friends is an English corpus based on the TV show Friends, containing 1,000 dialogues from seasons one to nine~\cite{chen2018emotionlines}. The 14,503 utterances from the 1,000 dialogues are categorised into seven classes. 
The annotators considered the context of the dialogue when labelling sentiments. 

\textbf {Mastodon}:
The Mastodon dataset~\cite{cerisara2018multi} consists of English posts from the Mastodon social media platform. While the dataset was initially designed for both sentiment recognition and dialogue act recognition, we only focus on the former. 

\textbf {MOSI}:
The Multimodal Opinion-level Sentiment Intensity (MOSI) dataset~\cite{zadeh2016multimodal} is a multimodal sentiment analysis dataset, including 2,199 opinion segments extracted from 93 videos. 
Each opinion segment received annotations on a sentiment spectrum ranging from highly negative to highly positive within the interval [-3, 3]. 

\textbf {CH-SIMS}:
CH-SIMS is a Chinese single- and multi-modal sentiment analysis dataset~\cite{yu2020ch}. It collected 2,281 video segments from movies, TV series, and a variety of shows. 
The sentiment annotation is divided into five categories. 

\textbf {M$^3$ED}:
Multi-modal Multi-scene Multi-label Emotional Dialogue (M$^3$ED) is the first multimodal emotional dialogue dataset in Chinese~\cite{DBLP:conf/acl/ZhaoZ0LJW022}. 
The dataset contains 990 dyadic emotional dialogues from 56 different TV series, including 9,082 turns and 24,449 utterances. 

\vspace{-5pt}
\subsection{Implementation Details}
We conducted emotion recognition tasks on six selected datasets to evaluate the effectiveness of two model adaptation methods on LLMs, \ie P-Tuning v2 and LoRA. For the SST dataset, we conducted both binary and five-class classification tasks. Three-class sentiment classification tasks, distinguishing among positive, neutral, and negative sentiments, were performed on the MOSI and the Mastodon dataset. The CH-SIMS dataset and the MOSI were used for a binary classification task: positive and negative. Finally, we implemented a seven-class emotion classification task on the Friends and M$^3$ED datasets. Moreover, for the purpose of performance comparison with specialised emotion recognition models, we have chosen recently published SOTA works that exhibit competitive performance on each selected datasets, separately, under strictly comparable conditions.

For each individual dataset, we designed three sets of comparative experiments: ChatGLM2 without adaptation, ChatGLM2 adapted with P-Tuning v2, and ChatGLM2 adapted with LoRA. 
During the inference with ChatGLM2, we utilise a ``prompt'' to acquire a response from it. The prompt should encompass both task-guiding sentences that necessitate emotion recognition. Our prompt is structured as follows: Classify the sentiment of the sentence to Emotion 1, Emotion 2, ... or Emotion k: $<$provide only one sentence from a test set$>$. The value of $k$ here is determined by the number of sentiment/emotion categories specific to the dataset. For example, the prompt is ``Classify the sentiment of the sentence to Positive, Negative or Neutral'' for MOSI as $k$ = 3. When adapting the model, we add a task-guiding sentence before each training sample to construct a complete prompt, and then present it into the model for learning. Note that, although the Mastodon, Friends, and M$^3$ED datasets are context-based, we treat them like other datasets, regardless of the context.

Our experiments were conducted on an NVIDIA GeForce RTX 3090 with 24GB of RAM, and the adaptation training and inference tasks were performed only on one single GPU.
For the adaptation training, we set the training batch size to 16 due to the constraints of GPU memory. Additionally, we set the prompt length to 32 for P-Tuning v2, and configured the rank of 8 for LoRA.
We employed the accuracy and macro F1 score as the primary metrics for performance evaluation. For the M$^3$ED dataset, we employed the weighted average F1 score to provide equitable comparisons with other research.

\begin{table}
\centering
\caption{Performance comparison between adapted models and SOTA works on the {\bf MOSI} datasets measured by accuracy (Acc) and macro-F1 (F1).}
\vspace{-5pt}
\label{tab:mosi}
\begin{tabular}{>{\centering\arraybackslash}p{3.5cm} c c | c c }
\toprule
\multicolumn{1}{c}{} & \multicolumn{2}{c}{MOSI-2} & \multicolumn{2}{c}{MOSI-3} \\
\cmidrule(lr){2-3} \cmidrule(l){4-5}
Model [\%] & Acc & F1 & Acc & F1 \\
\midrule
TFR-Net (2021)~\cite{DBLP:conf/mm/YuanLXY21} & 83.49 & - & - & - \\
CHFN (2022)~\cite{DBLP:conf/mm/GuoTDDK22} & 85.20& - & - & - \\
SeqSeq2Sent (2018)~\cite{DBLP:journals/corr/abs-1807-03915} & - & - & 77.00 & - \\
CTFN (2021)~\cite{DBLP:conf/acl/TangLJCZK20} & - & - & 80.79 & - \\
\midrule
ChatGLM2 & 84.12 & 84.12 & 77.26 & 58.19  \\
ChatGLM2 (P-Tuning) & 84.60 & 84.04 & 81.78 & \textbf{61.03}  \\
ChatGLM2 (LoRA) & \textbf{87.02} & \textbf{86.56} & \textbf{83.82} & 57.04 \\ 
\bottomrule
\end{tabular}
\end{table}

\begin{table}
\centering
\caption{Performance comparison between adapted models and SOTA works on the {\bf SST} datasets measured by accuracy (Acc) and macro-F1 (F1).}
\vspace{-5pt}
\label{tab:sst}
\begin{tabular}{c c c | c c }
\toprule
\multicolumn{1}{c}{} & \multicolumn{2}{c}{SST-2} & \multicolumn{2}{c}{SST-5}\\
\cmidrule(lr){2-3} \cmidrule(l){4-5} 
Model [\%] & Acc & F1 & Acc & F1 \\
\midrule
BT-TAPT (2021)~\cite{DBLP:journals/corr/abs-2107-10474}& 92.40 & - & - & - \\
SEMGraph-P (2022)~\cite{DBLP:conf/emnlp/WangLD0X22}& 94.23 & - & - & - \\
SentiLARE (2020)~\cite{DBLP:conf/emnlp/KeJLZH20}& - & - & 58.59 & - \\
SentiWSP (2022)~\cite{DBLP:conf/emnlp/0007L0LSZGGD22}& - & - & \textbf{59.32} & - \\
\midrule
ChatGLM2 & 82.33 & 82.33 & 30.09 & 25.82 \\
ChatGLM2 (P-Tuning) & 95.20 & 95.20 & 57.59 & \textbf{56.45} \\
ChatGLM2 (LoRA) & \textbf{95.69} & \textbf{95.69} & 54.45 & 52.51 \\
\bottomrule
\end{tabular}
\end{table}

\begin{table}[t]
\caption{Performance comparison between adapted models and SOTA works on the {\bf CH-SIMS} and {\bf Mastodon} datasets measured by accuracy (Acc) and macro-F1 (F1).}
\vspace{-5pt}
\label{tab:ch-sims}
\centering
\begin{tabular}{c c c | c c }
\toprule
\multicolumn{1}{c}{} & \multicolumn{2}{c}{CH-SIMS} & \multicolumn{2}{c}{Mastodon}\\
\cmidrule(lr){2-3} \cmidrule(l){4-5} 
Model [\%] & Acc & F1 & Acc & F1 \\
\midrule
MLF-DNN (2020)~\cite{yu2020ch}& 80.26 & - & - & - \\
DARER (2022)~\cite{DBLP:conf/acl/XingT22} & - & - & - & 59.59\\
\midrule
ChatGLM2 & 77.58 & 75.95 & 55.43 & 55.45 \\
ChatGLM2 (P-Tuning) & 82.47 & 81.12 & \textbf{67.25} & \textbf{67.23} \\
ChatGLM2 (LoRA) & \textbf{82.73} & \textbf{81.25} & 67.08 & 66.81 \\
\bottomrule
\end{tabular}
\end{table}

\begin{table}
\centering
\caption{Performance comparison on {\bf Friends} (first half) and {\bf M$^3$ED} (second half) in terms of accuracy (Acc), F1, and unweighted accuracy (UA). Note that, F1 indicates macro-F1 and weighted average F1 for Friends and M$^3$ED, respectively, for a fair performance comparison.}
\vspace{-5pt}
\label{tab:emotions}
\begin{tabular}
{p{4.5cm}<{\centering}p{0.7cm}<{\centering}p{0.7cm}<{\centering}p{0.7cm}<{\centering}}
\toprule 
  & \multicolumn{3}{c}{Friends} \\
  \cmidrule(l){2-4} 
 Model [\%] & Acc & F1 & UA \\
\midrule 
 BERT+SRL-GNN-8 (2020)~\cite{DBLP:conf/mm/HeatonS20}& 72.10 & - & 53.71\\
 XLNet+SRL-GNN-8 (2020)~\cite{DBLP:conf/mm/HeatonS20}& 72.82 & - & 53.41\\
 PRE-CODE (2020)~\cite{DBLP:conf/emnlp/JiaoLK20}& \textbf{81.30} & \textbf{65.90} & - \\
 \midrule 
 ChatGLM2 & 63.79 & 29.48 & 26.03 \\
 ChatGPT (P-Tuning) & 54.92 & 51.92 & \textbf{55.06} \\ 
 ChatGPT (LoRA) & 72.83 & 52.97 & 51.93 \\
 \midrule 
 \midrule 
  & \multicolumn{3}{c}{{M$^3$ED}} \\
  \cmidrule(l){2-4} 
 Model [\%]& Acc & F1 & UA \\
 \midrule 
 DialogueGCN (2019)~\cite{DBLP:conf/emnlp/GhosalMPCG19}& - & 46.09 & - \\
 DialogueRNN (2019)~\cite{DBLP:conf/aaai/MajumderPHMGC19}& - & 48.80 & -\\
 MDI (2022)~\cite{DBLP:conf/acl/ZhaoZ0LJW022}& - & \textbf{49.42} & - \\
 \midrule 
 ChatGLM2 & 45.68 & 30.52 & 16.82 \\
 ChatGLM2 (P-Tuning) & \textbf{45.75} & 37.31 & \textbf{28.64} \\ 
 ChatGLM2 (LoRA) & 42.54 & 33.31 & 23.59 \\
\bottomrule 
\end{tabular}
\end{table}

\vspace{-5pt}
\subsection{Results and Discussion}
To evaluate the effectiveness and the transferability of generalised LLMs in emotion recognition, we conducted extensive experiments on six publicly available datasets (see Section~\ref{sec:datasets}). Table~\ref{tab:mosi} to Table \ref{tab:emotions} 
present the results obtained from the ChatGLM2 with or without adaptation technologies on these six datasets, respectively. 
Besides, for each selected dataset, we offer SOTA performance from specialised models in the latest studies for comparison. 

First of all, we can see that ChatGLM2 performs competitively with these specialised models in many datasets, such as MOSI, CH-SIMS, and Mastodon. This finding is consistent with the one shown in our previous work but evaluated with other LLMs~\cite{DBLP:journals/corr/abs-2308-11578}. For the datasets of Friends and M$^3$ED, there is an obvious performance gap, which might be attributed to the lack of context information provided for ChatGLM2 for inference. 

Then, when comparing the performance of ChatGLM2 with or without adaptation, we can generally observe that the adapted large models, either by P-Tuning v2 or by LoRA, considerably outperform the non-adapted ones, both in binary and multi-class classification tasks. For instance, on the SST-5 dataset (cf.~Table 3), the P-Tuning v2 method performs the most substantial improvement. The accuracy and macro F1 scores increase from 30.09\,\% and 25.82\,\% to 57.59\,\% and 56.45\,\%, nearly doubling the performance before adaptation. This suggests that both P-Tuning v2 and LoRA algorithms work efficiently for the adaptation of LLMs in emotion recognition.  

Moreover, it can be seen that the adapted ChatGLM2s outperform other SOTA-specialised model in most cases, but vary depending on the complexity of the classification tasks. In simpler tasks like binary or three-class classification, adapted models often outperform SOTA-specialised models. Conversely, for tasks involving five or more categories (\eg Friends and M$3$ED), models with adaptation remain a substantial performance gap compared to SOTA works. This is largely due to the missing of context information for training and inference as aforementioned. Surprisingly, for the context-rich Mastodon dataset, even without considering context during adaptation, the adapted model exhibits superior performance compared to the SOTA works. This could be attributed to the relative simplicity of the three-class classification task or the dataset may have a less pronounced dependency on contextual information. 
Generally speaking, these observations indicate that the pretrained and generalised ChatGLM2 can efficiently transfer their knowledge to a specific domain without much training data and computation resources. 

Finally, in comparison with the two selected model adaptation methods, \ie P-Tuning v2 and LoRA, the latter outperforms in binary tasks, while the former demonstrates superior performance in ternary and multi-class tasks. Consequently, there is no consistent observation to definitively favour one method over the other, as the optimal adaptation approach varies across different datasets.

\vspace{-5pt}
\section{Conclusion}
In this paper, we focused on the capability of different model adaptation methods for Large Language Models (LLMs) in the field of emotion recognition. We investigate this by assessing the performance of the Chat General Language Model on six datasets using two adaptation techniques, \ie deep prompt tuning and low-rank adaptation. The experimental result shows that both adaptation methods perform exceptionally well in emotion recognition tasks, particularly for simple classification tasks that are without context. Compared to traditional specialised models, utilising the adapted LLMs for emotion recognition 
considerably
reduces the modelling efforts for researchers, and the computational resources required for adaptation are also accessible. This opens up brand-new possibilities for future emotion recognition systems.


\vfill\pagebreak

\bibliographystyle{IEEEtran}
\bibliography{refs}

\end{document}